\documentclass{article}
\usepackage{spconf}
\usepackage{amsmath}
\usepackage{amssymb}
\usepackage{graphicx}
\usepackage[utf8]{inputenc} 
\usepackage[T1]{fontenc}    
\usepackage{hyperref}       
\usepackage{url}            
\usepackage{booktabs}       
\usepackage{amsfonts}       
\usepackage{nicefrac}       
\usepackage{microtype}      
\usepackage{multirow}
\usepackage{dcolumn}
\usepackage{colortbl}
\usepackage{float}
\usepackage{enumerate}
\usepackage{cite}
\usepackage{import} 

\usepackage{tikz}
\usepackage{pgfplots}
\pgfplotsset{compat = 1.6}
\usetikzlibrary{calc}

\usepackage[belowskip=5pt,aboveskip=5pt]{caption}
\newcommand{\figscale}{0.75}

\title{Efficient Video and Audio processing with Loihi 2}
\name{Sumit Bam Shrestha$^{\star \dagger}$, Jonathan Timcheck, Paxon Frady, Leobardo Campos-Macias, Mike Davies }
\address{Intel Labs, Intel Corporation, Santa Clara, CA, 95054
USA \\ $^\dagger$sumit.bam.shrestha@intel.com }
\begin{document}
%
\maketitle
\begin{abstract}
  %
  %
  Loihi 2 is an asynchronous, brain-inspired research processor that generalizes several fundamental elements of neuromorphic architecture, such as stateful neuron models communicating with event-driven spikes, in order to address limitations of the first generation Loihi.
  Here we explore and characterize some of these generalizations, such as sigma-delta encapsulation, resonate-and-fire neurons, and integer-valued spikes, as applied to standard video, audio, and signal processing tasks. 
  We find that these new neuromorphic approaches can provide orders of magnitude gains in combined efficiency and latency (energy-delay-product) for feed-forward and convolutional neural networks applied to video, audio denoising, and spectral transforms compared to state-of-the-art solutions.



%
\end{abstract}

\section{Introduction}
\label{sec:intro}

In today's age of rapidly advancing Artificial Intelligence~(AI) capabilities with ever-growing, energy-hungry AI models, researchers are turning to novel computer architectures to unlock valuable efficiency improvements. One such promising novel architecture is neuromorphic computing, which aims to achieve vastly improved efficiency by applying computational principles inspired from the brain. 
While the GPUs, Tensor processors, and deep learning accelerators of today focus on dense matrix-based computation at a very high throughput, neuromorphic processors focus on sparse event-driven computation that minimizes activity and data movement.
Although neuromorphic processors are not yet mainstream commercial products, they have received increasing research and development focus in recent years, with an accelerating pace of progress. 
Some of the prominent neuromorphic hardware platforms include
Intel Loihi\cite{davies2018ieeemicro},
IBM TrueNorth\cite{merolla2014million}, 
SpiNNaker\cite{furber2020spinnaker}, 
Tianjic\cite{pei2019towards}, 
and SynSense Xylo\cite{bos2023sub}. 
All of these platforms focus on delivering efficient AI capabilities.

Neuromorphic processors use spiking neurons as their basic computational units. Historically, spiking neural networks~(SNNs) have been difficult to train, although recent progress in tools and methodologies\cite{neftci2019surrogate, shrestha2018neurips, wunderlich2021event} have enabled the training of deep SNNs at scale.
SNNs have been applied to image classification\cite{shrestha2018neurips, davies2021advancing}, gesture recognition\cite{shrestha2018neurips, davies2021advancing}, keyword spotting\cite{yin2021accurate}, visual-tactile sensing\cite{taunyazov2020event}, adaptive robotic arm control\cite{dewolf2023neuromorphic}, navigation\cite{tang2021deep}, and other tasks \cite{davies2021advancing}. Most of these SNN applications use a basic leaky integrate and fire~(LIF)\cite{gerstner2002spiking} neuron or some closely-related variant, and neurons communicate via binary spikes. Loihi 2 and other recent neuromorphic architectures have moved beyond simple LIF neurons with advanced new capabilities in order to expand application breadth and overcome algorithmic challenges. For example, Loihi 2 supports resonate-and-fire~(RF) neurons with complex-valued state and graded (integer-valued) spikes, which have shown their applicability in spectral processing of audio signals, optic flow estimation, keyword spotting, and automatic gain normalization\cite{frady2022efficient}. 
So far, however, there have been very few published results showing the value of these advanced neuron models and features from examples running and characterized on neuromorphic hardware.

\begin{figure}[!t]
    \centering
    \includegraphics[width=0.95\columnwidth]{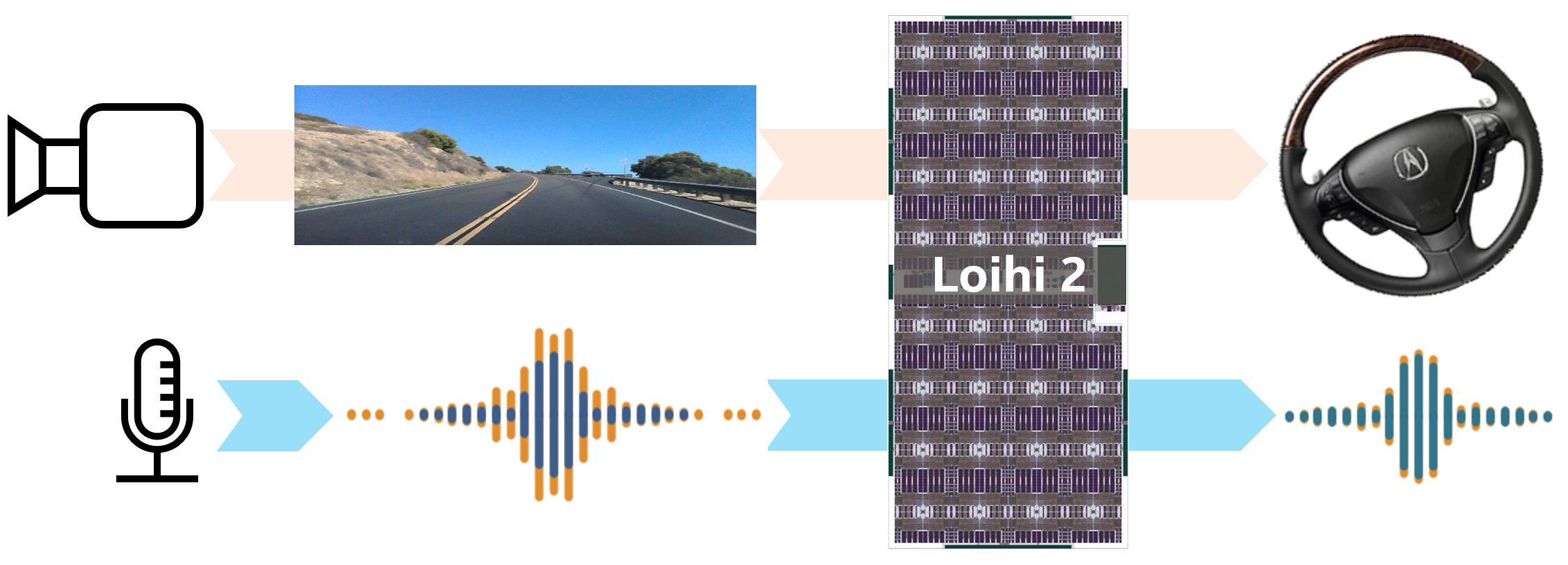}
    \caption{Efficient video and audio processing with Loihi 2.}
    \label{fig:loihi2solutions}
\end{figure}

In this paper, we first give a brief overview of Loihi 2 and the unique set of features it offers for efficient signal processing in Section~\ref{sec:loihi2}. In Section~\ref{sec:snn}, we discuss various families of spiking neurons that are available for processing video and audio data streams. 
Finally, in Section~\ref{sec:appl} we characterize and benchmark some representative examples that demonstrate efficient video and audio processing on Loihi 2 applied to practical problems.
\section{Loihi 2}
\label{sec:loihi2}



Loihi 2 is Intel's second-generation neuromorphic research chip featuring a massively parallel interconnect of fully asynchronous digital neuromorphic cores that communicate sparsely using spike messaging. Unlike its predecessor Loihi and many existing neuromorphic processors, Loihi 2 supports integer-valued spikes (called graded spikes), along with the usual binary spikes, at an insignificant additional energy cost. Each Loihi 2 neuromoprhic core, or neuro core, offers support for a wide variety of common synaptic connectivity topologies such as dense, sparse, convolution, as well as less common ones such as factorized or stochastic connections. Each neuro core can have up to 8190 neuron computation nodes which are programmable using a flexible microcode programmable neural engine to describe a multitude of neuron computation, dynamics, and spiking mechanisms.

\begin{figure}[!t]
    \centering
    \includegraphics[width=\columnwidth]{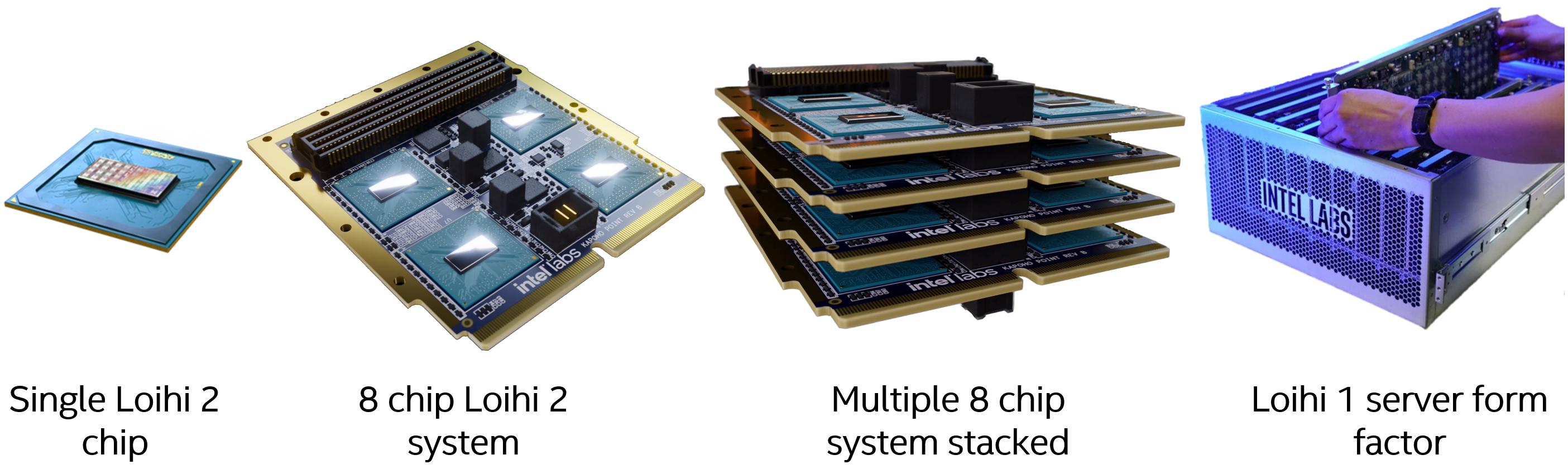}
    \caption{Range of Loihi 2 systems for signal processing at various scales.}
    \label{fig:loihi2system}
\end{figure}

A Loihi 2 chip consists of 126 or 128 neuro cores, depending on revision, which perform the bulk of neuromorphic computation; six embedded processor cores that enable direct interaction with the neuro core registers, management, and non-native communication; a dedicated spike I/O unit supporting off-chip communication over 10Gbps Ethernet at a throughput of up to 160 million spikes per second. In addition, the chip includes six asynchronous parallel interfaces that allow multiple Loihi 2 chips to be connected together in hierarchical mesh configurations ranging from one to thousands of chips. Various Loihi 2 form factors are shown in figure~\ref{fig:loihi2system}.


\section{Signal processing with generalized spiking neurons}
\label{sec:snn}




Spiking neurons are the fundamental computational units of the Loihi neuromorphic architecture.
Spiking neurons can be viewed as an upgraded version of the non-linear activation units used in conventional artificial neural networks (ANNs).
Historically, spiking neurons are mathematical models of biological neurons that communicate via unit-valued impulses that encode information exclusively in their spike times.
Recent neuromorphic chips such as Loihi 2 have generalized the concept of spiking neurons to include a wider variety of dynamics and communication in order to achieve greater efficiency and precision in digital circuit implementations. 
These generalizations preserve the fundamental properties of spiking neurons:
\begin{enumerate}
    \item All communication occurs in the form of asynchronous impulses called \textit{spikes}, with either single- or multi-level (graded) magnitudes. In Loihi 2, graded spikes carry up to 24 bits of integer magnitude. 
    \item A spiking neuron has one or more internal states that evolve over time according to the input the neuron receives and defined dynamics. Accordingly, spiking neurons are filters that transform their many inputs to a sparsely activated output signal.
\end{enumerate}


The most common spiking neuron in the neuromorphic computing literature is the biologically-inspired leaky integrate and fire (LIF) spiking neuron and similar variants, e.g., as implemented by Loihi 1\cite{davies2018ieeemicro}.
With the more flexible feature set of Loihi 2, it is now possible to deploy a broader range of spiking neuron models, such as sigma-delta encapsulated ReLU neurons and resonate-and-fire neurons.  This paper advances earlier simulation-based explorations of efficient signal processing using these spiking neuron models\cite{frady2022efficient} by characterizing several workload examples executing on Loihi 2 hardware.

\subsection{Sigma-Delta encapsulation}
\label{sec:snn:sdn}
Often in real-world problems, the inputs to and the outputs from ANN layers are temporally correlated. Hence, there is a significant amount of redundant computation being performed for each new frame or data point, but unfortunately, the non-linear activation functions of an ANN cannot take advantage of the data redundancy since the neurons are memoryless. Sigma-delta encapsulation takes advantage of temporal redundancy from the input data and between layers by communicating only the information that changed from the last timestep, thereby eliminating any unnecessary redundant computation.

The delta encoder is a simple frame difference operator that sends output graded spikes to subsequent layers only if the information exceeds its last communicated reference value, $x_\text{ref}[t]$, by a threshold $\vartheta$, i.e.,
\begin{equation}
\begin{aligned}
    s[t] &= (x[t] - x_\text{ref}[t - 1])\,\mathcal H(|x[t] - x_\text{ref}[t - 1]| - \vartheta) \\
    x_\text{ref}[t] &= x_\text{ref}[t - 1] + s[t].
\end{aligned}
\end{equation}
The delta encoder introduces sparsity in spike communication signals via its threshold mechanism.

\begin{figure}
    \centering
    \begin{tikzpicture}[thick, scale=\figscale, every node/.style={transform shape}]
    \begin{scope}[shift={(0, 0)}]
        \node (x) at (1, 0) {$x[t]$};
        \node (y) at (6.75, 0) {};
        \node at (6, 0.25) {$s[t]$};
        \node[draw] (z) at (2.5, 1) {$z^{-1}$};
        \node[draw, circle] (sum) at (2.5, 0) {+};
        \draw[->] (x) -- (sum);
        \draw[->] (z) -- (sum) node[pos=0.8, left] {$-$};
        \draw[->] (sum) ++ (2, 1) |- node[pos=0.3, right] {$x_\text{ref}[t]$} ++(-1, 0.5) -| (z);
        \draw[->] (sum) -- ++(1, 0);
        \draw (sum) ++ (1, -1) rectangle ++(2, 2);
        \draw[->] (sum) ++ (3, 0) -- (y);
        \draw[->] (sum) ++ (2, 0) ++(-0.75, 0) -- ++(1.5, 0);
        \draw[->] (sum) ++ (2, 0) ++(0, -0.75) -- ++(0, 1.5);
        \draw[thin] (sum) ++ (2, 0) ++(-0.25, 0) -- ++(0, -0.25) -- ++(-0.5, -0.5);
        \draw[thin] (sum) ++ (2, 0) ++( 0.25, 0) -- ++(0,  0.25) -- ++( 0.5,  0.5);
        \draw[thin, dotted] (sum) ++ (2, 0) ++(-0.25, -0.25) -- ++(0.5,  0.5);
        \node at (4, -1.3) {Delta encoding};
    \end{scope}

    \begin{scope}[shift={(7.75, 0)}]
        \node (x) at (0, 0) {};
        \node[text opacity=0.5] at (-0.5, 0) {SPARSE};
        \node (y) at (3.75, 0) {$x_\text{rec}[t]$};
        \node[draw] (z) at (2, 1) {$z^{-1}$};
        \node[draw, circle] (sum) at (1.25, 0) {+};
        \draw[->] (x) -- (sum);
        \draw[->] (sum) -- (y);
        \draw[->] (sum) ++ (1.5, 0) |- (z);
        \draw[->] (z) -| (sum);
        \node at (2, -1.3) {Sigma decoding};
    \end{scope}
\end{tikzpicture}
    \vspace{-0.5cm}
    \caption{Sparse communication of a temporally redundant signal using delta encoding and its corresponding reconstruction using sigma decoding.}
    \label{fig:sigma_delta}
\end{figure}
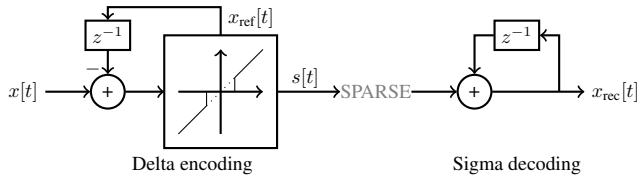

The sigma decoder on the receiving end performs a summation of the incoming signal to reconstruct the actual message:
\begin{align}
    x_\text{rec} &= x_\text{rec}[t - 1] + s[t]
\end{align}
The reconstruction of the message is not perfect because of the threshold mechanism. The value of the threshold controls the trade-off between perfect information transfer and the sparsity of the spikes being communicated. The sigma-delta mechanism is illustrated in figure~\ref{fig:sigma_delta}.

Sigma-delta encapsulation on the ReLU activation of ANNs is a straightforward way to perform efficient sparse computation on top of standard ANN processing using sigma-delta-ReLU units\cite{oconnor2016sigma}. The sigma-delta encapsulation, however, is more general and can be applied to any temporal processing neuron dynamics. We refer to Sigma-Delta Neural Networks as SDNNs.

\subsection{Resonate and Fire spiking neurons}
\label{sec:snn:rf}

In the biophysical mechanics of the brain, oscillatory dynamics are often observed. Oscillatory behaviors can be modeled by cross-coupling two state variables in the complex plane from the familiar Leaky Integrate and Fire decaying neural dynamics: given a complex-valued input $a[t]$, one can define the state dynamics of a complex-valued neuronal state $z[t]$ such that
\begin{align}
    z[t] = \lambda\,e^{\text i\omega_t} z[t - 1] + a,
    \label{eq:res}
\end{align}
where $\lambda \leq 1$ is the magnitude decay factor, $\omega_t$ is the frequency of oscillation, and an optional spike reset mechanism can apply (not written). Equation \eqref{eq:res} describes a Resonate and Fire (RF) neuron. Common spiking mechanisms of RF neurons are based on the phase and magnitude of the state. For instance, one can specify a spike to occur whenever the oscillation crosses some given half-plane (e.g., $\mathfrak{Im}(z) > \vartheta$), or the real part of the state $z[t]$ exceeds some given threshold \cite{izhikevich2001resonate, frady2022efficient}.

RF neurons not only have complex dynamics; they also take complex input coming from real and imaginary synapses. Thus, the computation using RF neurons gives rise to complex SNN topologies which are very expressive. Computation in the complex domain is a perfect fit for spectral analysis of audio signals \cite{frady2022efficient}. Additionally, RF neurons enable hyper-dimensional computing using Vector Symbolic Architecture~(VSA)\cite{kleyko2022vector} on Loihi 2.
\section{Applications}
\label{sec:appl}



In this section, we apply the novel neuromorphic neuron models described in Section~\ref{sec:snn} to video, audio, and signal processing tasks. All of the applications were implemented and mapped to Loihi 2 with the Lava\footnote{Lava is an open-source software framework for programming neuromorphic systems. It is available at \url{https://github.com/lava-nc/lava}.} software framework using the Lava-DL\footnote{Lava-DL is an open-source deep learning extension of Lava for training spiking neural networks, available at \url{https://github.com/lava-nc/lava-dl}.} library for training. 
We compare the applications' algorithmic performance, energy efficiency, latency, throughput, and energy-delay-product~(EDP) on Loihi 2 to the most comparable conventional workloads running on a standard commercially available edge computing platform, the NVIDIA Jetson Orin Nano, or an Intel Core i9-7920X CPU.


\begin{table*}[!ht]
    \centering
    \scriptsize
    \caption{Application performance comparison.}
    \label{table:results}
    \setlength\tabcolsep{3pt} 
    \begin{tabular}{l|c|l|r|r|r|r|r|r|r}
        \hline
        \multicolumn{1}{c|}{\multirow{3}{*}{\bf Network}} &
        \multicolumn{1}{c|}{\multirow{3}{*}{\bf Precision}} &
        \multicolumn{1}{c|}{\multirow{3}{*}{\bf Hardware}} &
        \multicolumn{1}{c|}{\multirow{3}{*}{\parbox{1.2cm}{\centering\bf Algorithmic quality}}} &
        \multicolumn{5}{c|}{\textbf{Hardware inference cost per sample}} &
        \multicolumn{1}{c}{\multirow{3}{*}{{\bf Param count}($\downarrow$)}} \\ \cline{5-9}
        &&& & \multicolumn{2}{c|}{Energy ($\downarrow$)} & \multicolumn{1}{c|}{Latency ($\downarrow$)} & \multicolumn{1}{c|}{Throughput ($\uparrow$)} & \multicolumn{1}{c|}{EDP ($\downarrow$)} & \\ \cline{5-9}
        &&&& \multicolumn{1}{c|}{Total (mJ)} &\multicolumn{1}{c|}{Dynamic (mJ)} &\multicolumn{1}{c|}{(ms)} &\multicolumn{1}{c|}{(samples/s)} &\multicolumn{1}{c|}{($\mu$Js)}\\\hline\hline

        &&& MSE (sq. rads) ($\downarrow$) &&&&&\\
        PilotNet SDNN            & int8 & Loihi 2$^\dagger$ (no IO)  & $0.035$ &  $0.09$ &   $0.05$ &  $1.21$ & $7403.80$ &   $0.11$ & $351,187$ \\
        PilotNet SDNN            & int8 & Loihi 2$^\dagger$ (IO limited) & $0.035$ &  $1.26$ &   $0.07$ & $65.41$ &  $137.60$ &  $82.54$ & $351,187$ \\
        PilotNet ANN (batch=1)   & fp32 & Jetson Orin Nano GPU$^\ddagger$ & $0.024$ & $21.94$ &  $10.12$ &  $5.77$ &  $173.19$ & $126.69$ & $351,187$ \\
        PilotNet ANN (batch=16)  & fp32 & Jetson Orin Nano GPU$^\ddagger$ & $0.024$ &  $6.14$ &   $3.72$ & $18.88$ &  $847.26$ &  $115.90$ & $351,187$ \\
        PilotNet ANN (batch=1)   & int8 & Jetson Orin Nano GPU$^\ddagger$ & $0.025$ & $13.72$ &   $6.70$ &  $3.43$ &  $291.48$ &  $47.08$ & $351,187$ \\
        PilotNet ANN (batch=16)  & int8 & Jetson Orin Nano GPU$^\ddagger$ & $0.025$ &  $5.31$ &   $2.89$ & $18.88$ &  $847.26$ &  $100.30$ & $351,187$ \\
        \hline\hline
        &&& SI-SNR (dB) ($\uparrow$) &&&&&&\\
        Intel NsSDNet$^*$\!\!    & int8 & Loihi 2 $^\dagger$ & $12.50$ &   $28.74$ &   $3.48$ & $32.04$ & $1.00$ &     $920.97$ &   $526,336$ \\
        Microsoft NsNet2$^*$\!\! & fp32 & Jetson Orin Nano GPU$^\ddagger$ & $11.89$ & $2143.35$ &  $95.35$ & $20.02$ & $1.00$ &  $42,909.90$ & $2,681,000$\\ \hline\hline
        &&& Correlation ($\uparrow$) &&&&&&\\
        RF STFT (5K events/s)    & int24 & Loihi 2$^\dagger$       & $0.94$ &   $0.34$ & $0.31$ & $0.82$ &  $59.39$ &   $0.28$ & $401$ \\
        STFT (500KB/s)           & fp32  & Core i9 CPU$^\mathsection$ & $0.98$ & $539.62$ &    -   & $0.43$ & $113.56$ & $232.02$ & - \\ \hline
        
        \multicolumn{10}{p{17.5cm}}{\scriptsize$^*$ The audio samples in the dataset \cite{timcheck2023intel} are 30 sec long but measurements are scaled for an effective 1 sec length for easier interpretation.
        $^\dagger$ The Loihi 2 networks were trained using Lava-dl 0.4.0. Workloads were characterized on an Oheo Gulch single-chip Loihi 2 system (N3C1 silicon for NDNS, N3B3 for others) running Lava 0.8.0 and Lava-Loihi 0.5.0 with an unreleased patch. 
        $^\ddagger$ ANN workloads were characterized on an NVIDIA Jetson Orin Nano 8GB 15W TDP running Jetpack 5.1.2, TensorRT 8.6.1, Torch-TensorRT 1.3.0. Energy values include CPU\_GPU\_CV and SOC components as reported by jtop. $^\mathsection$ CPU STFT characterization used the librosa 0.10.1 library running on an Intel Core i9-7920X with Ubuntu 20.04.6 LTS. Power measurements obtained with Intel SoC Watch 2021.2.  Performance results are based on testing as of September 2023 and may not reflect all publicly available security updates. Results may vary.
        } \\
    \end{tabular}
\end{table*}

\subsection{PilotNet: steering angle prediciton from dashboard RGB video}
\label{sec:appl:pilotnet}
PilotNet is an end-to-end deep learning application that predicts the steering angle of a car based on the input from a dashboard RGB camera\cite{bojarski2016end, bojarski2017explaining}. The network is a standard convolutional neural network consisting of five convolutional layers and four linear layers, each with ReLU activation. The network processes video input in the form of 200$\times$66 RGB frames.

The consecutive frames in the dashboard video have a high degree of temporal correlation and are therefore a prime candidate for sigma-delta sparsification. An SDNN with similar connectivity as the original PilotNet was trained using Lava-DL\footnote{The training code is available at \url{https://github.com/lava-nc/lava-dl}} with weights quantized to 8 bits and 16-bit activations, the maximum precision supported by Loihi 2. The SDNN implementation achieved a prediction mean square error of $0.035$ sq. radians compared to errors of $0.024$ and $0.025$ for the full-precision ANN model and an 8-bit quantized ANN model, respectively. Most of the sigma-delta model's increased error results from ReLU approximation error introduced by the sparse delta communication. The training makes use of Lava-DL's learnable threshold and additional sparsity specific loss in each layer to achieve the desired balance of prediction error and computation efficiency.  This results in 12$\times$ fewer synaptic operations and 11$\times$ fewer neuron activations compared to the ANN implementation.

We benchmarked two operational modes of the PilotNet SDNN on Loihi 2. The first mode injects input into the network at maximum throughput using a fixed data sequence stored in on-chip memory. In this mode, the SDNN can sustain a frame rate of 7.4K fps. The second mode streams the test dataset to the chip over an asynchronous parallel interface, which is currently constrained by software to a throughput of 137 samples/s.  The dynamic energy per inference of the two modes is similar, while the total energy of the latter mode is significantly higher since it is dominated by static power.  Likewise the latency of the IO-bottlenecked mode is over 50$\times$ higher since it is dominated by off-chip communication.

For comparison to the PilotNet on Jetson Orin Nano, we evaluated both non-batched and batched configurations of the ANN with the test set residing in system memory.  
The non-batched configuration is most comparable to Loihi 2 SDNN inference, which is fundamentally non-batched, and achieves minimum latency.  As batch size is increased, the ANN energy per inference improves up to batch size 16 at the expense of a linear degradation in latency.  In either case, the IO-unconstrained SDNN on  Loihi 2 achieves over 150$\times$ lower energy, 2.8$\times$ lower latency, and over 400$\times$ lower EDP.
Loihi 2 can achieve a 25$\times$ non-batched throughput advantage when not IO limited as a result of automatic layer-wise pipelining that the sigma-delta neuron model provides.


\subsection{Neuromorphic audio denoising}
\label{sec:appl:ndns}
Audio data is temporal and often requires efficient low-latency processing. This is especially true of a ubiquitous example, audio denoising.
In the release of the Intel Neuromorphic Deep Noise Suppression challenge\cite{timcheck2023intel},
we proposed the baseline Intel NsSDNet solution, an SDNN with other neuromorphic features such as axonal delays, that achieved superior noise reduction (as measured by SI-SNR) with nearly $10\times$ less computation than the Micosoft NsNet2\cite{braun2021towards}, the baseline ANN solution for the 2022 Microsoft DNS Challenge\cite{dubey2022icassp}. 
Advances in Lava software now allow us to characterize the Intel NsSDNet network on Loihi 2\footnote{The training script for the Intel NsSDNet is available at  \url{https://github.com/IntelLabs/IntelNeuromorphicDNSChallenge}.} in comparison to the Microsoft NsNet2 running on the Jetson Orin Nano.



We evaluated both denoising networks processing DNS challenge audio samples at a real-time sample rate of 16 kHz. The Loihi 2 implementation of the NsSDNet increases its power advantage to 74$\times$ compared to NsNet2 running on the Jetson Orin Nano platform. This increase compared to the difference in op counts is the result of the Orin Nano GPU architecture being unable to achieve peak efficiency for this real-time signal processing workload, in contrast to Loihi 2.





\subsection{Short-time Fourier transform using RF neurons}
\label{sec:appl:rfOF}

As described in \cite{frady2022efficient}, the short-time Fourier transform (STFT) of a signal can naturally be computed by resonate-and-fire neurons. 
The coefficients of the STFT are communicated through sparse spike-timing patterns.
The phase of the complex STFT coefficient is represented by the timing of the spikes \cite{frady2019robust}. On Loihi 2, the amplitudes of the complex coefficients are communicated by 24-bit integer graded spikes.

We measured the energy and latency of a bank of 200 resonate-and-fire neurons computing the STFT on Loihi 2 in comparison to a sliding-window STFT (window size 200 and hop length 1) on a CPU. 
The inputs were 8-16 second audio chirp signals, whose frequency increased over time. 
RF neuron resonant frequencies, $\omega$ in (\ref{eq:res}), were spaced evenly between 1Hz and 1kHz.
Note these units are parameterized based on a sampling rate of 4096 timesteps per second, but the algorithm was characterized in an accelerated execution mode, with Loihi 2 and the CPU each allowed to operate at its maximum throughput.

The power and latency results are shown in Table~\ref{table:results}. Not surprisingly, the CPU operating at 2.9 GHz gives 90\% higher STFT throughput and lower latency than the Loihi 2 RF neuron implementation. On the other hand, Loihi 2 is over 1500$\times$ more energy efficient than the CPU STFT and provides an overall EDP advantage of 
828$\times$. These gains come at the expense of an approximate algorithm. A reconstruction of the RF spectrogram correlates only to 0.94 with its input signal. Such a price is often acceptable for AI applications that do not require high fidelity waveform reconstructions.
\section{Conclusion}
\label{sec:concl}


In this paper, we have shown that a digital neuromorphic architectures such as Loihi 2 can provide significant gains in energy efficiency, latency, and even throughput for intelligent signal processing applications compared to conventional architectures.  In some cases these gains can exceed three orders of magnitude, albeit often at the cost of lower accuracy.  
A number of uniquely neuromorphic features enable these gains, such as stateful neurons with diverse dynamics, sparse yet graded spike communication, and an architecture that integrates memory and compute with highly granular parallelism to minimize data movement.
We believe further development will enable applications of neuromorphic computing that overcome power and latency constraints that currently limit the real-world, real-time deployment of AI capabilities.


\vfill\pagebreak

\small
\bibliographystyle{IEEEbib}
\bibliography{bibliography}

\end{document}